\documentclass[10pt,twocolumn,letterpaper]{article}

\usepackage{wacv}              %
\usepackage{graphicx}
\usepackage{amsmath}
\usepackage{amssymb}
\usepackage{booktabs}
\usepackage{xcolor}
\usepackage{multirow}
\usepackage{gensymb}
\usepackage{adjustbox}  %
\usepackage[margin=1in]{geometry}  %

\usepackage[pagebackref,breaklinks,colorlinks]{hyperref}

\usepackage[capitalize]{cleveref}
\crefname{section}{Sec.}{Secs.}
\Crefname{section}{Section}{Sections}
\Crefname{table}{Table}{Tables}
\crefname{table}{Tab.}{Tabs.}

\begin{document}

\title{Learning Keypoints for Multi-Agent Behavior Analysis using Self-Supervision}

\author{
Daniel Khalil $^{1}$
~~
Christina Liu$^{1}$
~~
Pietro Perona$^{1}$
~~
Jennifer J. Sun$^{1,2}$\thanks{Equal contribution.}
~~
Markus Marks$^{1*}$
\\\\
$^{1}$California Institute of Technology, $^{2}$Cornell University
}

\maketitle

\begin{abstract}

The study of social interactions and collective behaviors through multi-agent video analysis is crucial in biology.
While self-supervised keypoint discovery has emerged as a promising solution to reduce the need for manual keypoint annotations, existing methods often struggle with videos containing multiple interacting agents, especially those of the same species and color.
To address this, we introduce B-KinD-multi, a novel approach that leverages pre-trained video segmentation models to guide keypoint discovery in multi-agent scenarios. 
This eliminates the need for time-consuming manual annotations on new experimental settings and organisms. 
Extensive evaluations demonstrate improved keypoint regression and downstream behavioral classification in videos of flies, mice, and rats.  Furthermore, our method generalizes well to other species, including ants, bees, and humans, highlighting its potential for broad applications in automated keypoint annotation for multi-agent behavior analysis. 
Code available under: \href{https://danielpkhalil.github.io/B-KinD-Multi}{B-KinD-Multi}

\end{abstract}

\begin{figure}[t]
    \centering
    \includegraphics[clip, width=0.5\textwidth]{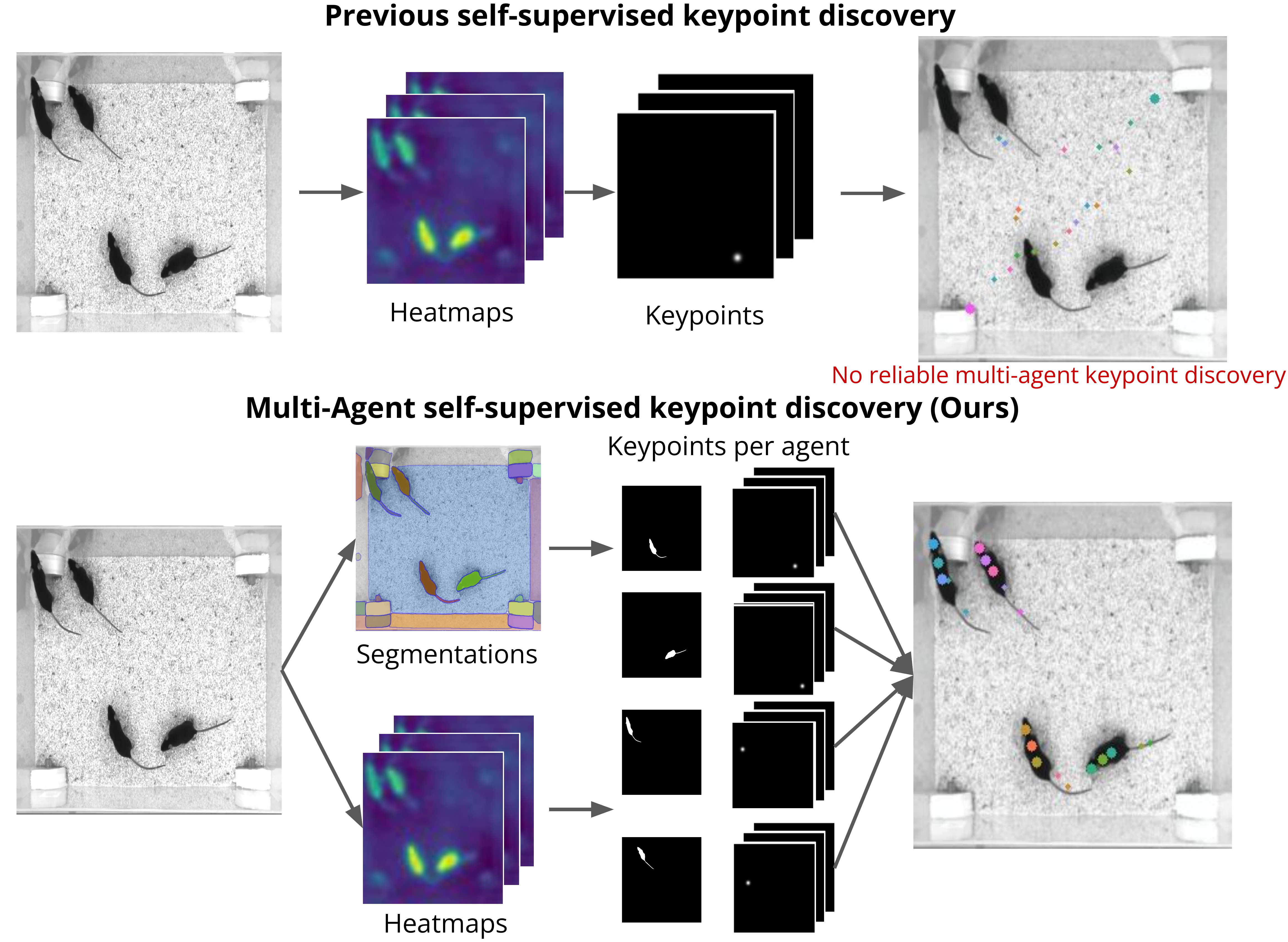}
    \caption{\textbf{Overview of B-KinD-multi:} We present B-KinD-multi a self-supervised keypoint discovery method for multiple agents. Previous methods struggle with discovering key points on multiple agents that look very similar. Our method addresses this issue by using video segmentation to mask feature maps during keypoint discovery. Our method enables accurate keypoint discovery and downstream behavioral classification.} 
    \label{fig:intro_fig}
\end{figure}

\begin{figure*}[t]
    \centering
    \includegraphics[clip, width=1.00\textwidth]{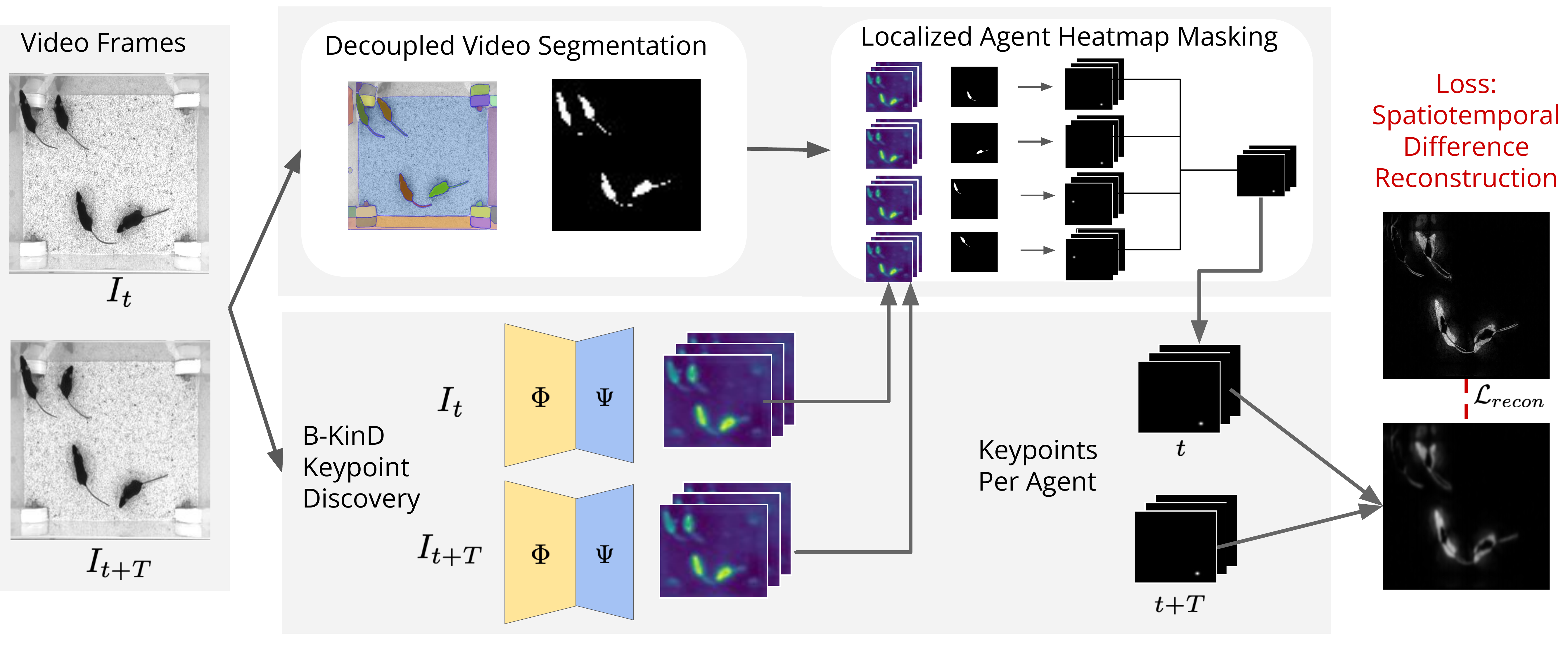}
    \caption{\textbf{Full overview of B-KinD-multi:} Video frames $I_t$ and $I_{t+T}$ at times $t$ and $t+T$ are processed through a video segmentation module to obtain agent masks. The raw frames are then fed into an appearance encoder $\Phi$, where appearance features are extracted from $I_t$, and a pose decoder $\Psi$. The agent masks are then separated, down-sampled, and used to produce keypoints per agent from both $I_t$ and $I_{t+T}$. These keypoints are then used as input to reconstruct the spatiotemporal difference computed from the two frames for each agent independently. This multi-agent approach allows for simultaneous keypoint discovery on multiple interacting subjects.}
    \label{fig:diagram}
\end{figure*}

\section{Introduction}
\label{sec:intro}

The analysis of behavior and interactions of multiple agents is critical for a wide range of applications, such as autonomous driving~\cite{chang2019argoverse,sun2020scalability}, video games~\cite{samvelyan2019starcraft,guss2019minerl}, sports analytics~\cite{yue2014learning,decroos2018automatic}, virtual worlds~\cite{hofmann2019minecraft}, and biology~\cite{anderson2014toward,datta2019computational}. Computer vision approaches provide tremendous potential for automatically analyzing multi-agent behavior across large-scale video datasets. 
A typical computer vision system for multi-agent behavior analysis consists of four core modules: agent detection, pose estimation and tracking, behavior classification, and future prediction. Our work focuses on the second stage: the challenging task of autonomous keypoint discovery and tracking, aiming to minimize the need for extensive human supervision.

Keypoints offer an efficient and effective way to analyze behavior, providing low-dimensional representations that capture essential features like body part distances, speed, and angles, while being computationally cheap to store\cite{sun2020task,nilsson2020simple}. Keypoints also offer fine-grained insights into specific movements, crucial for studying behaviors like gait in mice \cite{aljovic2022deep} or wing extensions in flies. 
Lastly, keypoints are very commonly used in many labs\cite{luxem2023open}.

Extracting keypoints is particularly critical in biology, where studying social behavior requires tracking and classifying the interactions between multiple animals~\cite{anderson2014toward}. 
Given the significant impact of social interactions on survival and reproduction, automated multi-agent behavior analysis promises to accelerate discoveries in ethology, neuroscience, pharmacology, and evolutionary biology~\cite{dell2014automated,markowitz2018striatum,stagkourakis2023anatomically,wiltschko2020revealing,hernandez2020framework,aljovic2022deep}.
Existing approaches to keypoint tracking often rely on supervised learning, necessitating large annotated datasets for training~\cite{dankert2009automated,branson2009high,kabra2013jaaba,eyjolfsdottir2014detecting,mathis2018deeplabcut,pereira2022sleap}. This requirement of laborious and time-consuming manual annotations is a bottleneck to the wider adoption of these methods, especially when dealing with new species or experimental setups.

Reducing human supervision, ideally to zero, has thus become a strategic priority for biology applications, especially in the analysis of social behavior~\cite{anderson2014toward,pereira2020quantifying}. It is, as it turns out, also an interesting challenge for computer vision researchers and a testbed for ideas in self-supervised and unsupervised video analysis.
A couple of self-supervised methods have appeared in the literature~\cite{jakab2018unsupervised, jakab2020self, zhang2018unsupervised}, with successful application to animal behavior analysis\cite{sun2022self}, including 3D pose estimation~\cite{sun2023bkind}. A caveat is that existing methods are designed for single-agent settings and, thus, struggle in social behavior analysis where multiple animals are present, particularly if these animals have the same coat color or appearance\cite{sun2022self}. This is because these models rely on tracking individual points across entire frames of videos, which becomes challenging when agents are visually similar and/or occlude each other.

To address this challenge, we propose a novel method for self-supervised keypoint discovery and tracking in videos containing multiple animals/agents. Our method integrates a decoupled video segmentation framework \cite{DEVA} into the keypoint discovery process through localized agent feature-map masking. 
We evaluate our method using videos containing multiple conspecifics, across three species of animals: flies, mice, and rats. We find that our method reduces both the error in keypoint localization, as well as downstream behavioral classification. Our work is the first to demonstrate a self-supervised multi-agent framework that learns keypoints without human annotations for new, unseen species.

Our contributions are as follows:
\begin{itemize}
    \item We introduce B-KinD-multi, a self-supervised keypoint discovery and tracking method for videos containing multiple animals of the same species and color.
    \item We quantitatively show improved keypoint regression in flies, mice, and rats and qualitatively demonstrate that our method also works on other species.
    \item We show how the reduced error in keypoint discovery leads to improved performance on downstream behavioral classification.
\end{itemize}

\section{Related Work}

\textbf{Pose Estimation}. 
Supervised approaches to pose estimation have achieved remarkable progress, ranging from human pose estimation \cite{toshev2014deeppose,newell2016stacked,wei2016convolutional,fang2017rmpe,cao2017realtime} to more specialized models for animal pose tracking like DeepLabCut \cite{mathis2018deeplabcut}. Recent approaches\cite{luxem2023open} have extended pose estimation frameworks to handle multiple animals\cite{lauer2022multi}, 3D poses\cite{gunel2019deepfly3d}, multiple species\cite{hayden2022automated,bala2020automated, doornweerd2021across} as well as in the wild\cite{wiltshire2023deepwild}. More recently, large-scale transformer models like UniPose \cite{yang2023unipose} have demonstrated impressive performance across many animal species. 
However, these models rely on substantial amounts of labeled data, limiting their adaptability to novel species or experimental settings without additional annotation. 
In contrast, our method focuses on enabling self-supervised keypoint discovery directly from video data, eliminating the reliance on manual annotations and facilitating broader applications in behavior analysis.

\textbf{Self-supervised keypoint discovery}. 
Representing behavior through keypoint trajectories improves computational efficiency and generalizability. However, obtaining keypoint annotations for different species in different settings is costly, and current methods do not readily generalize across experimental setups and species.
Self-supervised methods have emerged to address this, enabling direct keypoint discovery from video data. While initial work has focused on unsupervised part/keypoint discovery for humans \cite{jakab2018unsupervised, jakab2020self, zhang2018unsupervised}, recent work have extended these approaches to animal behavior analysis, particularly leveraging the controlled environment of laboratory settings with static backgrounds \cite{sun2022self, sun2023bkind}.
However, the inherent complexity of multi-agent videos, with occlusions and visually similar agents, presents significant challenges for existing self-supervised methods. Our work aims to bridge this gap by introducing a novel approach for multi-agent keypoint discovery.

\textbf{Behavior Quantification}. 
Quantifying behavior from animal videos is critical for many biological fields. 
Behavioral quantification is either done supervised\cite{sturman2020deep, segalin2020mouse, marks2022deep}, i.e. classifying individual frames into humanly defined behaviors or unsupervised, i.e. clustering frames into behavioral states\cite{berman2014mapping,wiltschko2015mapping,hsu2020b,luxem2020identifying}.
In both cases, the approach can use raw video frames\cite{marks2022deep, bohnslav2021deepethogram, wiltschko2015mapping, sun2023mabe22} or first process the video into trajectory data\cite{segalin2020mouse, sturman2020deep, weinreb2023keypoint}.
The extracted behavioral classes or states can be used to study neural activity\cite{markowitz2018striatum, stagkourakis2023anatomically}, functional or pharmacological perturbations~\cite{robie2017mapping,wiltschko2020revealing}.
Similar approaches have been used to study the behavior of animals in the wild, like in primatology\cite{bain2021automated, mannion2022multicomponent}.
Interactions of multiple animals are particularly important for understanding the evolutionary and neuronal basis of social interactions.
Several studies investigate the neural underpinnings of social behaviors such as aggression, mating, and courtship in mice and flies\cite{von2011neuronal,karigo2021distinct,stagkourakis2023anatomically}.

\section{Methods}

The goal of B-KinD-multi (Figure~\ref{fig:diagram}) is to discover semantically meaningful keypoints for multiple agents in behavioral videos without manual supervision. We extend the original B-KinD keypoint discovery framework~\cite{sun2022self} to robustly handle multiple agents of similar appearance, where the original keypoint discovery method struggles (\cref{fig:intro_fig}).

We use an encoder-decoder setup similar to previous methods~\cite{jakab2018unsupervised,ryou2021weakly,sun2022self}, but introduce a crucial segmentation step to isolate individual agent heatmaps and localize the keypoints per agent. As in the original B-KinD, we employ a reconstruction target based on spatiotemporal difference rather than full image reconstruction.
In behavioral videos with multiple agents, the camera is generally fixed with respect to the world, such that the background is largely stationary and the agents (\eg multiple mice moving in an enclosure) are the primary sources of motion. Spatiotemporal differences continue to provide a strong cue to infer location and movements of agents, but now we must differentiate between multiple moving entities.

Our approach first applies decoupled video segmentation~\cite{DEVA} to isolate individual agents in each frame. This segmentation step is critical for distinguishing between agents, especially of similar appearance. Following segmentation, we apply the modified B-KinD framework to each segmented region independently, allowing for keypoint discovery on a per-agent basis.

\subsection{Decoupled Video Segmentation}
For agent segmentation, we employ DEVA (Decoupled Video Segmentation)~\cite{DEVA}, a robust framework designed for video segmentation tasks. DEVA is particularly suitable for our multi-agent behavioral videos due to its ability to handle scenarios with limited task-specific data.

DEVA operates by combining two key components, an image segmentation model trained on the target task, providing task-specific image-level segmentation hypotheses, and a universal temporal propagation model trained on class-agnostic mask propagation datasets, which associates and propagates these hypotheses throughout the video. This  \textbf{Bi-Directional Propagation} is essential to DEVA's effectiveness and robustness in comparison to other methods such as ~\cite{SAM-Track}.

\subsubsection{In-clip Consensus}
The in-clip consensus step is crucial for denoising single-frame segmentations by considering a small clip of future frames. This process involves three main steps:

\textbf{Spatial Alignment:} Segmentations from different time steps are aligned to the current frame using the temporal propagation model. This alignment is crucial for computing correspondences between segments across frames.

\textbf{Representation:} After alignment, each segment becomes an object proposal for the current frame. These proposals are combined into a unified representation.

\textbf{Integer Programming:} An optimization problem is solved to select the best subset of proposals. The objective function balances two criteria:
\begin{enumerate}
    \item Selected proposals should be supported by other proposals (IoU $> 0.5$).
    \item Selected proposals should not significantly overlap with each other.
\end{enumerate}

This process results in a denoised consensus for the current frame, effectively filtering out noise and inconsistencies from single-frame segmentations.

\subsubsection{Merging Propagation and Consensus}
This step combines the propagated segmentation from past frames with the consensus from near-future frames, ensuring temporal consistency while incorporating new semantic information. The process involves:

\textbf{Association:} Segments from the propagated segmentation (past) and the consensus (near future) are associated based on their Intersection over Union (IoU).

\textbf{Fusion:} Associated segments are merged, while unassociated segments from both sources are preserved in the final output.

\textbf{Segment Deletion:} To manage computational costs, inactive segments (those that fail to associate with consensus segments for a specified number of consecutive times) are removed from memory.

This merging process allows B-KinD-multi to maintain temporal consistency while adapting to new information in the video stream.
For our B-KinD-multi framework, we utilize DEVA to generate agent masks $M_t^n$ and $M_{t+T}^n$ for each agent $n$ at time $t$ and $t+T$. These masks are crucial for isolating individual agents in subsequent keypoint discovery steps, allowing our model to handle multiple agents of similar appearance effectively.

\subsection{Self-supervised multi-agent keypoint discovery} 

Given a behavioral video containing multiple agents, our work aims to reconstruct regions of motion for each agent between a reference frame $I_t$ (the video frame at time $t$) and a future frame $I_{t+T}$ (the video frame $T$ timesteps later, for some set value of $T$). We accomplish this by first extracting appearance features and heatmaps, then segmenting individual agent heatmaps, and finally extracting keypoint locations ("geometry features") for each agent from both frames $I_t$ and $I_{t+T}$ (Figure~\ref{fig:diagram}).
We use an encoder-decoder architecture similar to the original B-KinD, with shared appearance encoder $\Phi$, geometry decoder $\Psi$, and reconstruction decoder $\psi$. The key difference lies in the introduction of an agent segmentation step and the per-agent processing of features.

\textbf{Individual Agent Mask Extraction} During training, the pair of frames $I_t$ and $I_{t+T}$ are first passed through an unsupervised segmentation algorithm to obtain agent masks $M_t$ and $M_{t+T}$ (in practice, we found text-prompted segmentation to be the most effective ~\cite{liu2023grounding, ren2024grounded}). Both $M_t$ and $M_{t+T}$ are split into individual agent masks $M_t^n$ and $M_{t+T}^n$ for each agent $n \in {1,\dots,N}$, where $N$ is the number of agents detected in the frame pair. These masks are used to isolate individual agents in subsequent processing steps. The pair of frames $I_t$ and $I_{t+T}$ are also fed to the appearance encoder $\Phi$ to generate appearance features, and those features are then fed into the geometry decoder $\Psi$ to generate geometry features. Similarly to \cite{sun2022self} the reference frame $I_t$ is used to generate both appearance and geometry representations, and the future frame $I_{t+T}$ is only used to generate a geometry representation. The appearance feature $h_a^t$ for frame $I_t$ are defined simply as the output of $\Phi$: $h_a^t = \Phi(I_t)$. The pose decoder $\Psi$ outputs $K$ raw heatmaps $\textbf{X}_k \in \mathbb{R}^2$, where $K$ is the number of keypoints. 

\textbf{Localized Agent Heatmap Masking: } Prior to masking, $N$ copies of the $K$ raw heatmaps are made so we have $X_{k,n}$ to replace $X_{k}$, and each $M_t^n$ and $M_{t+T}^n$ are down-sampled to match the dimension of $X_{k,n}$. Then, for each agent $n \in {1,\dots,N}$, a copy of the $K$ heatmaps are masked using the separated $N$ masks from $M_t^n$ and $M_{t+T}^n$. Specifically, for each agent $n$, we compute $KN$ masked heatmaps: $X_{k,n}^{mask} = X_{k} \odot M_t^n$ where $\odot$ denotes element-wise multiplication. A spatial softmax operation is then applied on each of the $K$ heatmap channel for each of the $N$ agents. With the extracted $p_{k,n}=(u_{k,n}, v_{k,n})$ locations for $k=\{1,\dots,K\}$ keypoints of $n=\{1,\dots,N\}$ agents from the spatial softmax, we define the geometry features $h_g^t$ to be a concatenation of 2D Gaussians centered at $(u_{k,n}, v_{k,n})$ with variance $\sigma$, resulting in $NK$ total keypoints.

Finally, the concatenation of the appearance feature $h_a^t$ and the geometry features $h_g^t$ and $h_g^{t+T}$ is fed to the decoder $\psi$ to reconstruct the learning objective $\hat{S}$ discussed in the next section: $\hat{S} = \psi(h_a^t, h_g^t, h_g^{t+T})$.

\subsection{Learning formulation}
\subsubsection{Spatiotemporal difference}
\label{sec:img_diff}
Our method for B-KinD-multi utilizes spatiotemporal differences as reconstruction targets, similar to the original B-KinD approach. Based on prior results from ~\cite{sun2022self}, we employ \textbf{Structural Similarity Index Measure} (SSIM))~\cite{Wang04imagequality} instead of absolute difference ($S_{\left|d\right|} = \left|I_{t+T} - I_t\right|$) or raw difference ($S_{d} = I_{t+T} - I_t$). This measure assesses the perceived quality of two images based on luminance, contrast, and structure features. For our multi-agent scenario, we apply SSIM locally on corresponding patches between $I_{t}$ and $I_{t+T}$ for each segmented agent to build a similarity map. The dissimilarity is then computed by negating this similarity map.

\subsubsection{Reconstruction loss}
We apply perceptual loss~\cite{Johnson2016Perceptual} for reconstructing the spatiotemporal difference $S$. This loss compares the L2 distance between features computed from a VGG network $\phi$~\cite{VGG14}. The reconstruction $\hat{S}$ and the target $S$ are fed into the VGG network, and mean squared error is applied to the features from the intermediate convolutional blocks:
\begin{align}
\mathcal{L}_{recon} = \left\Vert \phi(S(I_t,I_{t+T})) - \phi(\hat{S}(I_t,I_{t+T})) \right\Vert_2.
\end{align}

\subsubsection{Rotation equivariance loss}
To ensure semantic consistency of keypoints for agents that can move in various directions, we enforce rotation-equivariance. We apply the rotation equivariance loss on the generated heatmap for each agent, similar to the deformation equivariance in~\cite{Thewlis_2017_ICCV}.

Given reference image $I$ and the corresponding geometry bottleneck $h_g$, we rotate the geometry bottleneck to generate pseudo labels $h_g^{R\degree}$ for rotated input images $I^{R\degree}$ with degree $R=\{90\degree, 180\degree, 270\degree\}$. We apply mean squared error between the predicted geometry bottlenecks $\hat{h_g}$ from the rotated images and the generated pseudo labels $h_g$:
\begin{align}
    \mathcal{L}_{r} = \left\Vert h_g^{R\degree} - \hat{h}_g(I^{R\degree}) \right\Vert_2.
\end{align}

\subsubsection{Separation loss}
To prevent discovered keypoints from converging at the center of each agent's image, we apply a separation loss. This encourages keypoints to encode unique coordinates within each agent's segmented region. The separation loss for each agent $n$ is defined as:
\begin{align}
    \mathcal{L}_{s} = \sum_{i \neq j} \exp{\left( \frac{-(p_i - p_j)^2}{2\sigma_s^2} \right)}.
\end{align}

\subsubsection{Final objective}\label{sec:final_objective}
Our final loss function for B-KinD-multi combines the reconstruction loss $\mathcal{L}_{recon}$, rotation equivariance loss $\mathcal{L}r$, and separation loss $\mathcal{L}s$:
\begin{equation}
\mathcal{L} = \mathcal{L}_{\mathrm{recon}} + \mathbf{1}_{\mathrm{epoch} > n} \left( w_r \mathcal{L}_r + w_s \mathcal{L}_s \right)
\end{equation}

We employ curriculum learning~\cite{Bengio2009}, applying $\mathcal{L}_r$ and $\mathcal{L}_s$ only after the keypoints are consistently discovered from the semantic parts of each agent instance. We performed additional ablations, quantifying the impact of each of the losses (see supplement.

\subsection{Feature Extraction for Behavior Analysis}~\label{sec:behavior_quant}

B-KinD-multi, like its predecessor, uses discovered keypoints as input to behavior quantification modules. We extract additional features from the raw heatmaps, including confidence scores and shape information, which provide valuable insights into keypoint localization quality and body part configurations.
Given a raw heatmap $\textbf{X}_{i,k}$ for part $k$ of agent $n$, we compute a confidence score (defined as maximum value from the heatmap) and shape information (covariance matrix from the heatmap).

Using the normalized heatmap as a probability distribution, we calculate geometric features:
\begin{align}
    \sigma^2_x (\textbf{X}_{k,n}) &= \sum_{ij} (x_i - u_k)^2 \textbf{X}_{k,n}(i,j), \nonumber\\
    \sigma^2_y (\textbf{X}_{k,n}) &= \sum_{ij} (y_j - v_k)^2 \textbf{X}_{k,n}(i,j), \\
    \sigma^2_{xy} (\textbf{X}_{k,n}) &= \sum_{ij} (x_i - u_k) (y_j - v_k) \textbf{X}_{k,n}(i,j). \nonumber
\end{align}
These features, combined with the discovered keypoints, provide a comprehensive input for downstream behavior analysis tasks.

\begin{figure*}[t]
    \centering
    \begin{tabular}{@{}c@{\hspace{0.01\linewidth}}c@{\hspace{0.01\linewidth}}c@{\hspace{0.01\linewidth}}|@{\hspace{0.01\linewidth}}c@{\hspace{0.01\linewidth}}c@{\hspace{0.01\linewidth}}c@{}}
        \multicolumn{3}{c}{B-KinD} & \multicolumn{3}{c}{B-KinD-Multi} \\
        \adjustbox{valign=m}{\includegraphics[width=0.12\textwidth]{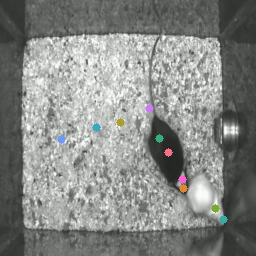}} & 
        \adjustbox{valign=m}{\includegraphics[width=0.12\textwidth]{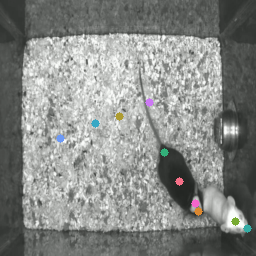}} & 
        \adjustbox{valign=m}{\includegraphics[width=0.12\textwidth]{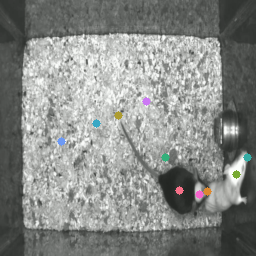}} & 
        \adjustbox{valign=m}{\includegraphics[width=0.12\textwidth]{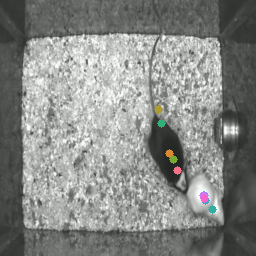}} & 
        \adjustbox{valign=m}{\includegraphics[width=0.12\textwidth]{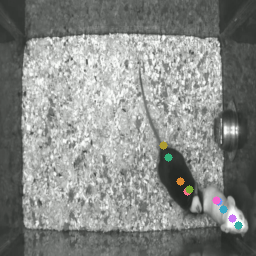}} & 
        \adjustbox{valign=m}{\includegraphics[width=0.12\textwidth]{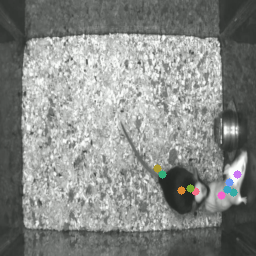}} \\
        \adjustbox{valign=m}{\includegraphics[width=0.12\textwidth]{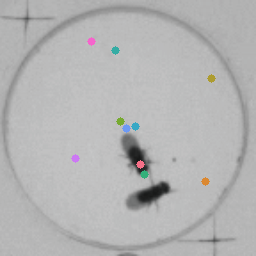}} & 
        \adjustbox{valign=m}{\includegraphics[width=0.12\textwidth]{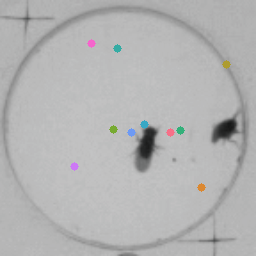}} & 
        \adjustbox{valign=m}{\includegraphics[width=0.12\textwidth]{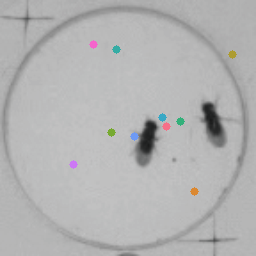}} & 
        \adjustbox{valign=m}{\includegraphics[width=0.12\textwidth]{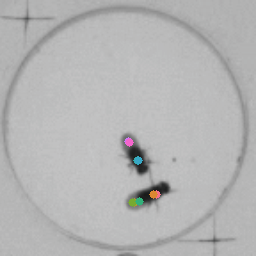}} & 
        \adjustbox{valign=m}{\includegraphics[width=0.12\textwidth]{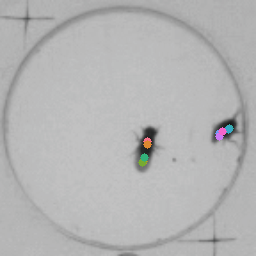}} & 
        \adjustbox{valign=m}{\includegraphics[width=0.12\textwidth]{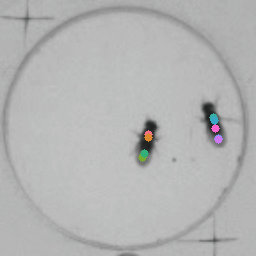}} \\
        \adjustbox{valign=m}{\includegraphics[width=0.12\textwidth]{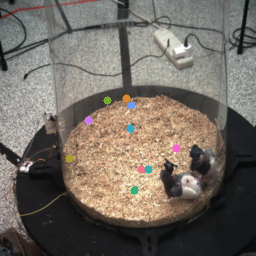}} & 
        \adjustbox{valign=m}{\includegraphics[width=0.12\textwidth]{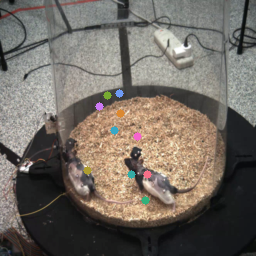}} & 
        \adjustbox{valign=m}{\includegraphics[width=0.12\textwidth]{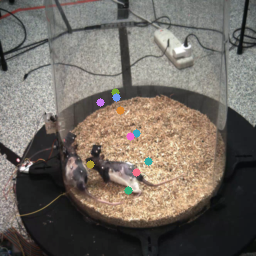}} & 
        \adjustbox{valign=m}{\includegraphics[width=0.12\textwidth]{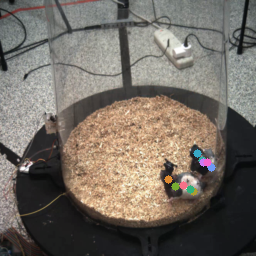}} & 
        \adjustbox{valign=m}{\includegraphics[width=0.12\textwidth]{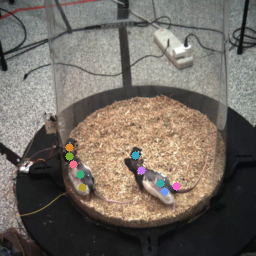}} & 
        \adjustbox{valign=m}{\includegraphics[width=0.12\textwidth]{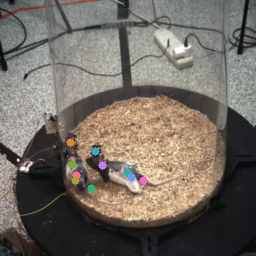}} \\
    \end{tabular}
    \caption{\textbf{Qualitative Comparisons between B-KinD and B-KinD-multi.} We compare the keypoint discovery performance qualitatively between B-KinD and B-KinD-multi on Fly v. Fly, CalMS21 and PAIR-R24. We find that for each dataset B-KinD-multi performs much better than B-KinD. The difference is bigger for datasets where the agents have the same color (Fly v. Fly and PAIR-R24) compared to when they are different (CalMS21).}
    \label{fig:qualitative_comparison}
\end{figure*}

\section{Results}

\subsection{Experimental Setup}

\subsubsection{Datasets}
\textbf{CalMS21}: This comprehensive dataset focuses on mouse interactions, featuring video and trajectory data of paired mice. Expert annotations cover three key behaviors: sniffing, attacking, and mounting. Our training utilized the 507,000-frame train split, excluding videos with miniscope cables. For behavior classification, we followed the protocol established by Sun et al. (2021), using the entire training split for classifier training and the 262,000-frame test split for evaluation. The high-resolution frames (1024x570 pixels) capture detailed mouse interactions.

\textbf{MARS-Pose}: We employed a subset of this mouse behavior dataset to assess our model's ability to predict human-annotated keypoints. Our experiments used varying training set sizes (10, 50, 100, and 500 images) with a consistent test set of 1,500 images, allowing us to evaluate performance across different levels of supervision.

\textbf{Fly vs. Fly}: This dataset presents interactions between fruit fly pairs, with frame-by-frame expert annotations. We focused on the Aggression videos, which comprise 1,229,000 frames for training and 322,000 for testing. Despite the lower resolution (144x144 pixels), these videos capture intricate fly behaviors. Following previous work, we evaluated performance on behaviors with substantial representation in the training set (over 1,000 frames), specifically: lunging, wing threats, and tussling.

\textbf{PAIR-R24M}: This dataset, designed for multi-animal 3D pose estimation, provides an extensive collection of dyadic rat interactions. It comprises 24.3 million frames of RGB video coupled with 3D ground-truth motion capture data, featuring 18 distinct rat pairs captured from 24 different viewpoints. We adapted this 3D dataset for 2D keypoint regression task by projecting the 3D coordinates onto 2D planes depending on the camera view and calibration.

\subsubsection{Training and Evaluation Methodology}
We trained B-KinD-multi using the comprehensive objective outlined in the methods. During the training process, images were rescaled to 256 × 256 pixels, with a temporal window T of approximately 0.2 seconds for most datasets. Unless otherwise specified, all experiments utilized the full set of keypoints discovered by B-KinD-multi with SSIM reconstruction. We used 10 keypoints for mice and flies, and 12 keypoints for rats in the PAIR-R24M dataset. Training was conducted on the specified train split of each dataset. Post-training, we extracted keypoints for various evaluations based on the available labels in each dataset. Our evaluation approach differed across datasets:

Behavior Classification: For CalMS21 and Fly vs. Fly datasets, we followed the original B-KinD methodology. We trained the 1D Convolutional Network benchmark model provided by~\cite{sun2021multi} using our B-KinD-multi keypoints from the train split of each dataset. Evaluation was performed using mean average precision (MAP), weighted equally over all behaviors of interest.

Keypoint Regression: We extended our keypoint regression evaluation to CalMS21, Fly vs. Fly, and PAIR-R24M datasets, adopting an approach similar to that used for Human3.6M in the original B-KinD paper. For these datasets, we employed a linear regressor without a bias term, following the evaluation setup from previous works ~\cite{ZhangKptDisc18,Lorenz19}. Evaluation was based on mean squared error (\%-MSE) normalized by image size.

\begin{table}
  \begin{center}
\scalebox{0.7}{
    \begin{tabular}{lcccc}
        \toprule[0.2em]
        CalMS21 & Pose & Conf & Cov & MAP\\
        \toprule[0.2em]
        \multicolumn{5}{c}{\textbf{\textit{Fully supervised}}} \\
        \multirow{3}{*}{MARS $\dagger$ ~\cite{segalin2020mouse}} & \checkmark & & & $.856 \pm .010$ \\
        & \checkmark & \checkmark & & $.874 \pm .003$ \\
        & \checkmark & \checkmark & \checkmark & $.880 \pm .005$ \\
        \bottomrule[0.1em]
        \multicolumn{5}{c}{\textbf{\textit{Self-supervised}}} \\   
        Jakab et al.~\cite{jakab2018unsupervised} & \checkmark & &  & $.186 \pm .008$\\
        \hline
        \multirow{3}{*}{Image Recon.} & \checkmark & & & $.182 \pm .007$ \\
        & \checkmark & \checkmark & & $.184 \pm .006$ \\
        & \checkmark & \checkmark & \checkmark & $.165 \pm .012$ \\
        \hline
        \multirow{3}{*}{Image Recon. bbox$\dagger$} & \checkmark & & & $.819 \pm .008$ \\
        & \checkmark & \checkmark & & $.812 \pm .006$ \\
        & \checkmark & \checkmark & \checkmark & $.812 \pm .010$ \\ \hline
        \multirow{3}{*}{B-KinD} & \checkmark & & & $.814 \pm .007$ \\
        & \checkmark & \checkmark & & $.857 \pm .005$ \\
        & \checkmark & \checkmark & \checkmark & $.852 \pm .013$ \\  \hline
        \multirow{3}{*}{Ours} & \checkmark & & & $.834 \pm .010$ \\
        & \checkmark & \checkmark & & $.862 \pm .011$ \\
        & \checkmark & \checkmark & \checkmark & $.864 \pm .029$ \\   
        \bottomrule[0.1em]
    \end{tabular}}
  \caption{\textbf{Behavior Classification Results on CalMS21}. ``Ours" represents classifiers using input keypoints from our discovered keypoints. ``conf" represents using the confidence score, and ``cov" represents values from the covariance matrix of the heatmap. $\dagger$ refers to models that require bounding box inputs before keypoint estimation. Mean and std dev from 5 classifier runs are shown. }
  \label{tab:mouse_behavior}
  \end{center}\vspace{-0.4cm}
\end{table}

\subsection{Behavior Classification Results}\label{sec:behavior_res}
We evaluated the effectiveness of B-KinD-multi for behavior classification on both the CalMS21 and Fly vs. Fly datasets, comparing our results with the original B-KinD and other baseline methods.

\textbf{CalMS21 Behavior Classification}.
Table~\ref{tab:mouse_behavior} presents the behavior classification results on the CalMS21 dataset. Our B-KinD-multi method performs better than the original B-KinD across all input configurations. When using pose, confidence, and covariance as input, B-KinD-multi achieves a mean average precision (MAP) of $.864 \pm .029$, surpassing B-KinD ($.852 \pm .013$) and approaching the performance of the fully supervised MARS method ($.880 \pm .005$).
Notably, B-KinD-multi outperforms other self-supervised methods, including those that require bounding box inputs. Our method's performance without using bounding boxes ($.834 \pm .010$ with pose only) is superior to the image reconstruction with bounding boxes ($.819 \pm .008$), highlighting the effectiveness of our multi-agent approach in capturing relevant behavioral features.
The inclusion of confidence scores and covariance information from the heatmaps continues to improve performance, as observed in the original B-KinD. This suggests that our multi-agent extension preserves the valuable information captured in these additional features.

\textbf{Fly Behavior Classification}.
For the Fly vs. Fly dataset (Table~\ref{tab:fly_behavior}), B-KinD-multi again shows improvement over the original B-KinD, achieving a MAP of $.768 \pm .031$ compared to B-KinD's $.727 \pm .022$. This performance is closer to the hand-crafted features of FlyTracker ($.809 \pm .013$) and surpasses the image reconstruction with bounding boxes ($.750 \pm .020$).
As with the original B-KinD, we computed generic features (speed, acceleration, distance, and angle) from the discovered keypoints without assuming keypoint identity. The improved performance of B-KinD-multi suggests that our method is better at capturing the subtle interactions and behaviors of multiple agents, even in the challenging scenario of fly interactions.
These results demonstrate that B-KinD-multi not only preserves the strengths of the original B-KinD in behavior classification but also enhances its performance through improved multi-agent keypoint discovery, especially in cases of agents with similar appearance. The consistent improvement across both datasets indicates that our method is effective in capturing the complexities of multi-agent interactions, leading to more accurate behavior classification in various species and experimental settings.

\begin{table}
  \begin{center}
  \vspace{-0.05in}
\scalebox{0.7}{
    \begin{tabular}{lc}
        \toprule[0.2em]
        Fly &  MAP\\
        \toprule[0.2em]
        \multicolumn{2}{c}{\textbf{\textit{Hand-crafted features}}} \\
        FlyTracker~\cite{eyjolfsdottir2014detecting} & $.809 \pm .013$ \\
        \bottomrule[0.1em]
        \multicolumn{2}{c}{\textbf{\textit{Self-supervised + generic features}}} \\    
        Image Recon. & $.500 \pm .024$  \\
        Image Recon. bbox$\dagger$ & $.750 \pm .020$ \\
        B-KinD & $.727 \pm .022$ \\
        Ours & $.768 \pm .031$ \\
        \bottomrule[0.1em]
    \end{tabular}}
  \vspace{-0.05in}    
  \caption{\textbf{Behavior Classification Results on Fly vs. Fly}. ``FlyTracker" represents classifiers using hand-crafted inputs from~\cite{eyjolfsdottir2014detecting}. The self-supervised keypoints all use the same ``generic features" computed on all keypoints: speed, acceleration, distance, and angle. $\dagger$ refers to models that require bounding box inputs before keypoint estimation. Mean and std dev from 5 classifier runs are shown. }\vspace{-0.6cm}
  \label{tab:fly_behavior}
  \end{center}
\end{table}

\subsection{Keypoint regression results}

We evaluated the pose estimation performance of B-KinD-multi on three datasets: CalMS21, Fly vs. Fly, and PAIR-R24M. Our results demonstrate significant improvements in keypoint regression accuracy compared to the original B-KinD method across all datasets. \cref{fig:qualitative_comparison} shows qualitatively the differences in performance across all three datasets.

\textbf{CalMS21 Keypoint Regression}.
Table~\ref{tab:merged_regression} presents the keypoint regression results on the CalMS21 dataset. B-KinD-multi achieves a mean squared error (MSE) of $.693\% \pm .061\%$, substantially outperforming the original B-KinD (1.681\% ± .122\%). This represents a 58.8\% reduction in error, indicating that our multi-agent approach significantly improves keypoint localization accuracy in mouse interactions.

\textbf{Fly vs. Fly Keypoint Regression}.
For the Fly vs. Fly dataset (Table~\ref{tab:merged_regression}), B-KinD-multi demonstrates even more dramatic improvement. Our method achieves an MSE of $.115\% \pm .032\%$, compared to B-KinD's 1.179\% $\pm$ .048\%. This represents a 90.2\% reduction in error, showcasing the effectiveness of our approach in handling the challenging task of keypoint regression in small, fast-moving subjects like flies.

\textbf{PAIR-R24M Keypoint Regression}.
Table~\ref{tab:merged_regression} shows the results on the PAIR-R24M dataset, which was not included in the original B-KinD evaluation. On this new multi-agent rat interaction dataset, B-KinD-multi achieves an MSE of $.168\% \pm .027\%$, significantly outperforming B-KinD ($.927\% \pm .060\%$). This 81.9\% error reduction demonstrates that our method generalizes well to new datasets and species, maintaining its effectiveness in multi-agent scenarios.
These results consistently show that B-KinD-multi substantially improves keypoint regression accuracy across diverse datasets. The significant reductions in error rates (58.8\% for CalMS21, 90.2\% for Fly vs. Fly, and 81.9\% for PAIR-R24M) highlight the effectiveness of our multi-agent approach in capturing accurate keypoint locations.

\begin{figure}[t]
    \centering
    \begin{tabular}{@{}c@{\hspace{0.02\linewidth}}c@{}}
        \includegraphics[width=0.4\linewidth]{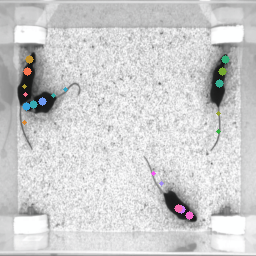} & 
        \includegraphics[width=0.4\linewidth]{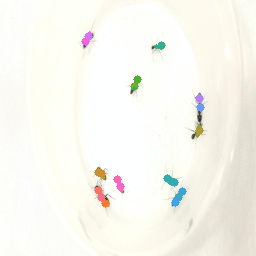} \\
        \makebox[0.4\linewidth]{\centering 4 Mice (black)} & 
        \makebox[0.4\linewidth]{\centering 10 Ants} \\
        \includegraphics[width=0.4\linewidth]{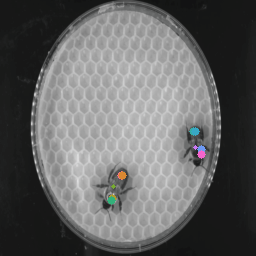} & 
        \includegraphics[width=0.4\linewidth]{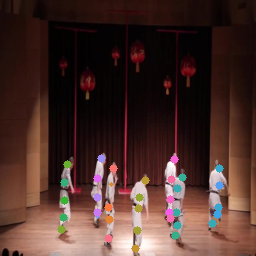} \\
        \makebox[0.4\linewidth]{\centering 2 Bees} & 
        \makebox[0.4\linewidth]{\centering 7 Humans} \\
    \end{tabular}
    \caption{\textbf{Qualitative results of B-KinD-multi on other datasets.} We demonstrate qualitatively the performance of B-KinD-multi on other multi-agent datasets. From left to right: 4 mice, 10 ants, 2 bees, 7 humans.}
    \label{fig:additional_qualtiative}
\end{figure}

\begin{table}
\begin{center}
\vspace{-0.05in}
\scalebox{0.7}{
\begin{tabular}{lccc}
\toprule[0.2em]
Method & Dataset & \%-MSE \\
\midrule[0.2em]
\multicolumn{3}{c}{\textbf{\textit{Self-supervised}}} \\
\cmidrule(lr){1-3}
B-KinD & CalMS21 & $ 1.681 \pm .122$ \\
Ours & CalMS21 & $ .693 \pm .061$ \\
\midrule
\multicolumn{3}{c}{} \\
B-KinD & Fly & $ 1.179 \pm .048$ \\
Ours & Fly & $ .115 \pm .032$ \\
\midrule
\multicolumn{3}{c}{} \\
B-KinD & PAIR-R24M & $.927 \pm .060$ \\
Ours & PAIR-R24M & $.168 \pm .027$ \\
\bottomrule[0.1em]
\end{tabular}
}
\vspace{-0.05in}
\caption{\textbf{Keypoint Regression Results}. Comparison of keypoint regression performance on CalMS21, Fly, and PAIR-R24M datasets. Results are shown in mean squared error (\%-MSE) normalized by image size for two self-supervised keypoint discovery methods: B-KinD and our proposed approach. Lower values indicate better performance. Reported values are mean and standard deviation over 5 runs.}
\label{tab:merged_regression}
\end{center}
\end{table}

\subsection{Effect of Video Segmentation type on keypoint discovery}

To evaluate the impact of different video segmentation approaches on B-KinD-multi's performance, we conducted an ablation study comparing two methods for creating agent masks: a frame-by-frame approach based on the SAM Track \cite{cheng2023segment}, and the DEVA \cite{DEVA} segmentation method which employs a bidirectional framework for improved mask creation.

We show the effect of these segmentation methods on both behavior classification and keypoint regression tasks (see supplement).

The results demonstrate that the choice of video segmentation method significantly impacts B-KinD-multi's performance on all datasets. These results highlight the importance of high-quality, temporally consistent agent masks in multi-agent keypoint discovery and behavior analysis. 

These improvements come at an additional computational cost, which we quantified in the supplement.

\subsection{Other qualitative results}

For evaluating B-KinD-multi quantitatively, we were limited by the annotated multi-agent video-based datasets available. To demonstrate the power of our method beyond those datasets, we include some qualitative results on other datasets and species. Specifically, \cref{fig:additional_qualtiative} shows 4 new settings: 4 black mice, 10 ants, 2 bees, and 7 humans dancing.

\section{Discussion and Conclusions}

B-KinD-multi is a novel method for self-supervised keypoint discovery and tracking in video containing multiple animals / agents.  Our method builds upon B-KinD, a framework for self-supervised keypoint discovery in video containing single animals or animals with different appearances.
In B-KinD-multi decoupled video segmentation masks the features of individual animals in the keypoint discovery model, thus enabling per-animal keypoint detection and tracking.

In experiments on multi-animal video across three species, we compare the accuracy of B-KinD-multi with that of B-KinD. We find that B-KinD-multi selects more informative keypoints, and reduces keypoint regression error, particularly for agents of the same color. We also find that reduced keypoint regression error translates to an increased downstream behavioral classification performance.

A limitation of B-KinD-multi is that it assumes static backgrounds. Static backgrounds are common in lab environments, as well as in video taken by camera-traps in the wild. Video taken by hand-held devices, for example, when studying the behavior of apes in their natural environment, does not satisfy this assumption and will thus require a more sophisticated method.  
Future work to extend our capabilities to handle dynamic backgrounds would broaden the applicability of our method to these real-world scenarios.

We find our method robust in terms of the number of agents. Further research is needed to see the limitations concerning the number of agents. Different types of occlusions should also be evaluated.

\section*{Acknowledgements}
Pietro Perona and Markus Marks were supported by the National Institutes of Health (NIH R01 MH123612A), the Caltech Chen Institute (Neuroscience Research Grant Award), and the Simons Foundation (NC-GB-CULM-00002953-02). Daniel Khalil was supported by the Caltech SURF program.

{\small
\bibliographystyle{ieee_fullname}
\bibliography{egbib}

\begin{thebibliography}{10}\itemsep=-1pt

\bibitem{aljovic2022deep}
Almir Aljovic, Shuqing Zhao, Maryam Chahin, Clara de~la Rosa, Valerie Van~Steenbergen, Martin Kerschensteiner, and Florence~M Bareyre.
\newblock A deep learning-based toolbox for automated limb motion analysis (alma) in murine models of neurological disorders.
\newblock {\em Communications Biology}, 5(1):131, 2022.

\bibitem{anderson2014toward}
David~J Anderson and Pietro Perona.
\newblock Toward a science of computational ethology.
\newblock {\em Neuron}, 84(1):18--31, 2014.

\bibitem{bain2021automated}
Max Bain, Arsha Nagrani, Daniel Schofield, Sophie Berdugo, Joana Bessa, Jake Owen, Kimberley~J Hockings, Tetsuro Matsuzawa, Misato Hayashi, Dora Biro, et~al.
\newblock Automated audiovisual behavior recognition in wild primates.
\newblock {\em Science advances}, 7(46):eabi4883, 2021.

\bibitem{bala2020automated}
Praneet~C Bala, Benjamin~R Eisenreich, Seng Bum~Michael Yoo, Benjamin~Y Hayden, Hyun~Soo Park, and Jan Zimmermann.
\newblock Automated markerless pose estimation in freely moving macaques with openmonkeystudio.
\newblock {\em Nature communications}, 11(1):4560, 2020.

\bibitem{Bengio2009}
Yoshua Bengio, J{\'e}r\^{o}me Louradour, Ronan Collobert, and Jason Weston.
\newblock Curriculum learning.
\newblock In {\em ICML}, 2009.

\bibitem{berman2014mapping}
Gordon~J Berman, Daniel~M Choi, William Bialek, and Joshua~W Shaevitz.
\newblock Mapping the stereotyped behaviour of freely moving fruit flies.
\newblock {\em Journal of The Royal Society Interface}, 11(99):20140672, 2014.

\bibitem{bohnslav2021deepethogram}
James~P Bohnslav, Nivanthika~K Wimalasena, Kelsey~J Clausing, Yu~Y Dai, David~A Yarmolinsky, Tom{\'a}s Cruz, Adam~D Kashlan, M~Eugenia Chiappe, Lauren~L Orefice, Clifford~J Woolf, et~al.
\newblock Deepethogram, a machine learning pipeline for supervised behavior classification from raw pixels.
\newblock {\em Elife}, 10:e63377, 2021.

\bibitem{branson2009high}
Kristin Branson, Alice~A Robie, John Bender, Pietro Perona, and Michael~H Dickinson.
\newblock High-throughput ethomics in large groups of drosophila.
\newblock {\em Nature methods}, 6(6):451--457, 2009.

\bibitem{cao2017realtime}
Zhe Cao, Tomas Simon, Shih-En Wei, and Yaser Sheikh.
\newblock Realtime multi-person 2d pose estimation using part affinity fields.
\newblock In {\em Proceedings of the IEEE conference on computer vision and pattern recognition}, pages 7291--7299, 2017.

\bibitem{chang2019argoverse}
Ming-Fang Chang, John Lambert, Patsorn Sangkloy, Jagjeet Singh, Slawomir Bak, Andrew Hartnett, De Wang, Peter Carr, Simon Lucey, Deva Ramanan, et~al.
\newblock Argoverse: 3d tracking and forecasting with rich maps.
\newblock In {\em Proceedings of the IEEE Conference on Computer Vision and Pattern Recognition}, pages 8748--8757, 2019.

\bibitem{DEVA}
Ho~Kei Cheng, Seoung~Wug Oh, Brian Price, Alexander Schwing, and Joon-Young Lee.
\newblock Tracking anything with decoupled video segmentation.
\newblock In {\em Proceedings of the IEEE/CVF International Conference on Computer Vision}, pages 1316--1326, 2023.

\bibitem{SAM-Track}
Yangming Cheng, Liulei Li, Yuanyou Xu, Xiaodi Li, Zongxin Yang, Wenguan Wang, and Yi Yang.
\newblock Segment and track anything.
\newblock {\em arXiv preprint arXiv:2305.06558}, 2023.

\bibitem{cheng2023segment}
Yangming Cheng, Liulei Li, Yuanyou Xu, Xiaodi Li, Zongxin Yang, Wenguan Wang, and Yi Yang.
\newblock Segment and track anything.
\newblock {\em arXiv preprint arXiv:2305.06558}, 2023.

\bibitem{dankert2009automated}
Heiko Dankert, Liming Wang, Eric~D Hoopfer, David~J Anderson, and Pietro Perona.
\newblock Automated monitoring and analysis of social behavior in drosophila.
\newblock {\em Nature methods}, 6(4):297--303, 2009.

\bibitem{datta2019computational}
Sandeep~Robert Datta, David~J Anderson, Kristin Branson, Pietro Perona, and Andrew Leifer.
\newblock Computational neuroethology: a call to action.
\newblock {\em Neuron}, 104(1):11--24, 2019.

\bibitem{decroos2018automatic}
Tom Decroos, Jan Van~Haaren, and Jesse Davis.
\newblock Automatic discovery of tactics in spatio-temporal soccer match data.
\newblock In {\em Proceedings of the 24th acm sigkdd international conference on knowledge discovery \& data mining}, pages 223--232, 2018.

\bibitem{dell2014automated}
Anthony~I Dell, John~A Bender, Kristin Branson, Iain~D Couzin, Gonzalo~G de Polavieja, Lucas~PJJ Noldus, Alfonso P{\'e}rez-Escudero, Pietro Perona, Andrew~D Straw, Martin Wikelski, et~al.
\newblock Automated image-based tracking and its application in ecology.
\newblock {\em Trends in ecology \& evolution}, 29(7):417--428, 2014.

\bibitem{doornweerd2021across}
Jan~Erik Doornweerd, Gert Kootstra, Roel~F Veerkamp, Esther~D Ellen, Jerine~AJ van~der Eijk, Thijs van~de Straat, and Aniek~C Bouwman.
\newblock Across-species pose estimation in poultry based on images using deep learning.
\newblock {\em Frontiers in Animal Science}, 2:791290, 2021.

\bibitem{eyjolfsdottir2014detecting}
Eyrun Eyjolfsdottir, Steve Branson, Xavier~P Burgos-Artizzu, Eric~D Hoopfer, Jonathan Schor, David~J Anderson, and Pietro Perona.
\newblock Detecting social actions of fruit flies.
\newblock In {\em European Conference on Computer Vision}, pages 772--787. Springer, 2014.

\bibitem{fang2017rmpe}
Hao-Shu Fang, Shuqin Xie, Yu-Wing Tai, and Cewu Lu.
\newblock Rmpe: Regional multi-person pose estimation.
\newblock In {\em Proceedings of the IEEE international conference on computer vision}, pages 2334--2343, 2017.

\bibitem{gunel2019deepfly3d}
Semih G{\"u}nel, Helge Rhodin, Daniel Morales, Jo{\~a}o Campagnolo, Pavan Ramdya, and Pascal Fua.
\newblock Deepfly3d, a deep learning-based approach for 3d limb and appendage tracking in tethered, adult drosophila.
\newblock {\em Elife}, 8:e48571, 2019.

\bibitem{guss2019minerl}
William~H Guss, Brandon Houghton, Nicholay Topin, Phillip Wang, Cayden Codel, Manuela Veloso, and Ruslan Salakhutdinov.
\newblock Minerl: A large-scale dataset of minecraft demonstrations.
\newblock {\em arXiv preprint arXiv:1907.13440}, 2019.

\bibitem{hayden2022automated}
Benjamin~Y Hayden, Hyun~Soo Park, and Jan Zimmermann.
\newblock Automated pose estimation in primates.
\newblock {\em American journal of primatology}, 84(10):e23348, 2022.

\bibitem{hernandez2020framework}
Dami{\'a}n~G Hern{\'a}ndez, Catalina Rivera, Jessica Cande, Baohua Zhou, David~L Stern, and Gordon~J Berman.
\newblock A framework for studying behavioral evolution by reconstructing ancestral repertoires.
\newblock {\em arXiv preprint arXiv:2007.09689}, 2020.

\bibitem{hofmann2019minecraft}
Katja Hofmann.
\newblock Minecraft as ai playground and laboratory.
\newblock In {\em Proceedings of the Annual Symposium on Computer-Human Interaction in Play}, pages 1--1, 2019.

\bibitem{hsu2020b}
Alexander~I Hsu and Eric~A Yttri.
\newblock B-soid: An open source unsupervised algorithm for discovery of spontaneous behaviors.
\newblock {\em bioRxiv}, page 770271, 2020.

\bibitem{jakab2018unsupervised}
Tomas Jakab, Ankush Gupta, Hakan Bilen, and Andrea Vedaldi.
\newblock Unsupervised learning of object landmarks through conditional image generation.
\newblock {\em Advances in neural information processing systems}, 31, 2018.

\bibitem{jakab2020self}
Tomas Jakab, Ankush Gupta, Hakan Bilen, and Andrea Vedaldi.
\newblock Self-supervised learning of interpretable keypoints from unlabelled videos.
\newblock In {\em Proceedings of the IEEE/CVF Conference on Computer Vision and Pattern Recognition}, pages 8787--8797, 2020.

\bibitem{Johnson2016Perceptual}
Justin Johnson, Alexandre Alahi, and Li Fei-Fei.
\newblock Perceptual losses for real-time style transfer and super-resolution.
\newblock In {\em European Conference on Computer Vision}, 2016.

\bibitem{kabra2013jaaba}
Mayank Kabra, Alice~A Robie, Marta Rivera-Alba, Steven Branson, and Kristin Branson.
\newblock Jaaba: interactive machine learning for automatic annotation of animal behavior.
\newblock {\em Nature methods}, 10(1):64, 2013.

\bibitem{karigo2021distinct}
Tomomi Karigo, Ann Kennedy, Bin Yang, Mengyu Liu, Derek Tai, Iman~A Wahle, and David~J Anderson.
\newblock Distinct hypothalamic control of same-and opposite-sex mounting behaviour in mice.
\newblock {\em Nature}, 589(7841):258--263, 2021.

\bibitem{lauer2022multi}
Jessy Lauer, Mu Zhou, Shaokai Ye, William Menegas, Steffen Schneider, Tanmay Nath, Mohammed~Mostafizur Rahman, Valentina Di~Santo, Daniel Soberanes, Guoping Feng, et~al.
\newblock Multi-animal pose estimation, identification and tracking with deeplabcut.
\newblock {\em Nature Methods}, 19(4):496--504, 2022.

\bibitem{liu2023grounding}
Shilong Liu, Zhaoyang Zeng, Tianhe Ren, Feng Li, Hao Zhang, Jie Yang, Chunyuan Li, Jianwei Yang, Hang Su, Jun Zhu, et~al.
\newblock Grounding dino: Marrying dino with grounded pre-training for open-set object detection.
\newblock {\em arXiv preprint arXiv:2303.05499}, 2023.

\bibitem{Lorenz19}
Dominik Lorenz, Leonard Bereska, Timo Milbich, and Bj{\"o}rn Ommer.
\newblock Unsupervised part-based disentangling of object shape and appearance.
\newblock In {\em CVPR}, 2019.

\bibitem{luxem2020identifying}
Kevin Luxem, Falko Fuhrmann, Johannes K{\"u}rsch, Stefan Remy, and Pavol Bauer.
\newblock Identifying behavioral structure from deep variational embeddings of animal motion.
\newblock {\em bioRxiv}, 2020.

\bibitem{luxem2023open}
Kevin Luxem, Jennifer~J Sun, Sean~P Bradley, Keerthi Krishnan, Eric Yttri, Jan Zimmermann, Talmo~D Pereira, and Mark Laubach.
\newblock Open-source tools for behavioral video analysis: Setup, methods, and best practices.
\newblock {\em Elife}, 12:e79305, 2023.

\bibitem{mannion2022multicomponent}
Kelly~Ray Mannion, Elizabeth~F Ballare, Markus Marks, and Thibaud Gruber.
\newblock A multicomponent approach to studying cultural propensities during foraging in the wild.
\newblock {\em The Journal of Animal Ecology}, 2022.

\bibitem{markowitz2018striatum}
Jeffrey~E Markowitz, Winthrop~F Gillis, Celia~C Beron, Shay~Q Neufeld, Keiramarie Robertson, Neha~D Bhagat, Ralph~E Peterson, Emalee Peterson, Minsuk Hyun, Scott~W Linderman, et~al.
\newblock The striatum organizes 3d behavior via moment-to-moment action selection.
\newblock {\em Cell}, 174(1):44--58, 2018.

\bibitem{marks2022deep}
Markus Marks, Qiuhan Jin, Oliver Sturman, Lukas von Ziegler, Sepp Kollmorgen, Wolfger von~der Behrens, Valerio Mante, Johannes Bohacek, and Mehmet~Fatih Yanik.
\newblock Deep-learning-based identification, tracking, pose estimation and behaviour classification of interacting primates and mice in complex environments.
\newblock {\em Nature machine intelligence}, 4(4):331--340, 2022.

\bibitem{mathis2018deeplabcut}
Alexander Mathis, Pranav Mamidanna, Kevin~M Cury, Taiga Abe, Venkatesh~N Murthy, Mackenzie~Weygandt Mathis, and Matthias Bethge.
\newblock Deeplabcut: markerless pose estimation of user-defined body parts with deep learning.
\newblock {\em Nature neuroscience}, 21(9):1281--1289, 2018.

\bibitem{newell2016stacked}
Alejandro Newell, Kai Yang, and Jia Deng.
\newblock Stacked hourglass networks for human pose estimation.
\newblock In {\em European conference on computer vision}, pages 483--499. Springer, 2016.

\bibitem{nilsson2020simple}
Simon~RO Nilsson, Nastacia~L Goodwin, Jia~J Choong, Sophia Hwang, Hayden~R Wright, Zane Norville, Xiaoyu Tong, Dayu Lin, Brandon~S Bentzley, Neir Eshel, et~al.
\newblock Simple behavioral analysis (simba): an open source toolkit for computer classification of complex social behaviors in experimental animals.
\newblock {\em BioRxiv}, 2020.

\bibitem{pereira2020quantifying}
Talmo~D Pereira, Joshua~W Shaevitz, and Mala Murthy.
\newblock Quantifying behavior to understand the brain.
\newblock {\em Nature neuroscience}, pages 1--13, 2020.

\bibitem{pereira2022sleap}
Talmo~D Pereira, Nathaniel Tabris, Arie Matsliah, David~M Turner, Junyu Li, Shruthi Ravindranath, Eleni~S Papadoyannis, Edna Normand, David~S Deutsch, Z~Yan Wang, et~al.
\newblock Sleap: A deep learning system for multi-animal pose tracking.
\newblock {\em Nature methods}, 19(4):486--495, 2022.

\bibitem{ren2024grounded}
Tianhe Ren, Shilong Liu, Ailing Zeng, Jing Lin, Kunchang Li, He Cao, Jiayu Chen, Xinyu Huang, Yukang Chen, Feng Yan, Zhaoyang Zeng, Hao Zhang, Feng Li, Jie Yang, Hongyang Li, Qing Jiang, and Lei Zhang.
\newblock Grounded sam: Assembling open-world models for diverse visual tasks, 2024.

\bibitem{robie2017mapping}
Alice~A Robie, Jonathan Hirokawa, Austin~W Edwards, Lowell~A Umayam, Allen Lee, Mary~L Phillips, Gwyneth~M Card, Wyatt Korff, Gerald~M Rubin, Julie~H Simpson, et~al.
\newblock Mapping the neural substrates of behavior.
\newblock {\em Cell}, 170(2):393--406, 2017.

\bibitem{ryou2021weakly}
Serim Ryou and Pietro Perona.
\newblock Weakly supervised keypoint discovery.
\newblock {\em CoRR}, abs/2109.13423, 2021.

\bibitem{samvelyan2019starcraft}
Mikayel Samvelyan, Tabish Rashid, Christian~Schroeder De~Witt, Gregory Farquhar, Nantas Nardelli, Tim~GJ Rudner, Chia-Man Hung, Philip~HS Torr, Jakob Foerster, and Shimon Whiteson.
\newblock The starcraft multi-agent challenge.
\newblock {\em arXiv preprint arXiv:1902.04043}, 2019.

\bibitem{segalin2020mouse}
Cristina Segalin, Jalani Williams, Tomomi Karigo, May Hui, Moriel Zelikowsky, Jennifer~J Sun, Pietro Perona, David~J Anderson, and Ann Kennedy.
\newblock The mouse action recognition system (mars): a software pipeline for automated analysis of social behaviors in mice.
\newblock {\em bioRxiv}, 2020.

\bibitem{VGG14}
Karen Simonyan and Andrew Zisserman.
\newblock Very deep convolutional networks for large-scale image recognition.
\newblock {\em CoRR}, abs/1409.1556, 2014.

\bibitem{stagkourakis2023anatomically}
Stefanos Stagkourakis, Giada Spigolon, Markus Marks, Michael Feyder, Joseph Kim, Pietro Perona, Marius Pachitariu, and David~J Anderson.
\newblock Anatomically distributed neural representations of instincts in the hypothalamus.
\newblock {\em bioRxiv}, 2023.

\bibitem{sturman2020deep}
Oliver Sturman, Lukas von Ziegler, Christa Schl{\"a}ppi, Furkan Akyol, Mattia Privitera, Daria Slominski, Christina Grimm, Laetitia Thieren, Valerio Zerbi, Benjamin Grewe, et~al.
\newblock Deep learning-based behavioral analysis reaches human accuracy and is capable of outperforming commercial solutions.
\newblock {\em Neuropsychopharmacology}, 45(11):1942--1952, 2020.

\bibitem{sun2023bkind}
Jennifer~J Sun, Lili Karashchuk, Amil Dravid, Serim Ryou, Sonia Fereidooni, John~C Tuthill, Aggelos Katsaggelos, Bingni~W Brunton, Georgia Gkioxari, Ann Kennedy, et~al.
\newblock Bkind-3d: self-supervised 3d keypoint discovery from multi-view videos.
\newblock In {\em Proceedings of the IEEE/CVF Conference on Computer Vision and Pattern Recognition}, pages 9001--9010, 2023.

\bibitem{sun2021multi}
Jennifer~J Sun, Tomomi Karigo, Dipam Chakraborty, Sharada~P Mohanty, Benjamin Wild, Quan Sun, Chen Chen, David~J Anderson, Pietro Perona, Yisong Yue, et~al.
\newblock The multi-agent behavior dataset: Mouse dyadic social interactions.
\newblock {\em arXiv preprint arXiv:2104.02710}, 2021.

\bibitem{sun2020task}
Jennifer~J Sun, Ann Kennedy, Eric Zhan, David~J Anderson, Yisong Yue, and Pietro Perona.
\newblock Task programming: Learning data efficient behavior representations.
\newblock In {\em Proceedings of the IEEE/CVF Conference on Computer Vision and Pattern Recognition}, pages 2876--2885, 2021.

\bibitem{sun2023mabe22}
Jennifer~J Sun, Markus Marks, Andrew~Wesley Ulmer, Dipam Chakraborty, Brian Geuther, Edward Hayes, Heng Jia, Vivek Kumar, Sebastian Oleszko, Zachary Partridge, et~al.
\newblock Mabe22: A multi-species multi-task benchmark for learned representations of behavior.
\newblock In {\em International Conference on Machine Learning}, pages 32936--32990. PMLR, 2023.

\bibitem{sun2022self}
Jennifer~J Sun, Serim Ryou, Roni~H Goldshmid, Brandon Weissbourd, John~O Dabiri, David~J Anderson, Ann Kennedy, Yisong Yue, and Pietro Perona.
\newblock Self-supervised keypoint discovery in behavioral videos.
\newblock In {\em Proceedings of the IEEE/CVF Conference on Computer Vision and Pattern Recognition}, pages 2171--2180, 2022.

\bibitem{sun2020scalability}
Pei Sun, Henrik Kretzschmar, Xerxes Dotiwalla, Aurelien Chouard, Vijaysai Patnaik, Paul Tsui, James Guo, Yin Zhou, Yuning Chai, Benjamin Caine, et~al.
\newblock Scalability in perception for autonomous driving: Waymo open dataset.
\newblock In {\em Proceedings of the IEEE/CVF Conference on Computer Vision and Pattern Recognition}, pages 2446--2454, 2020.

\bibitem{Thewlis_2017_ICCV}
James Thewlis, Hakan Bilen, and Andrea Vedaldi.
\newblock Unsupervised learning of object landmarks by factorized spatial embeddings.
\newblock In {\em The IEEE International Conference on Computer Vision (ICCV)}, Oct 2017.

\bibitem{toshev2014deeppose}
Alexander Toshev and Christian Szegedy.
\newblock Deeppose: Human pose estimation via deep neural networks.
\newblock In {\em Proceedings of the IEEE conference on computer vision and pattern recognition}, pages 1653--1660, 2014.

\bibitem{von2011neuronal}
Anne~C Von~Philipsborn, Tianxiao Liu, Y~Yu Jai, Christopher Masser, Salil~S Bidaye, and Barry~J Dickson.
\newblock Neuronal control of drosophila courtship song.
\newblock {\em Neuron}, 69(3):509--522, 2011.

\bibitem{Wang04imagequality}
Zhou Wang, Alan~C. Bovik, Hamid~R. Sheikh, and Eero~P. Simoncelli.
\newblock Image quality assessment: From error visibility to structural similarity.
\newblock {\em IEEE TRANSACTIONS ON IMAGE PROCESSING}, 13(4):600--612, 2004.

\bibitem{wei2016convolutional}
Shih-En Wei, Varun Ramakrishna, Takeo Kanade, and Yaser Sheikh.
\newblock Convolutional pose machines.
\newblock In {\em Proceedings of the IEEE Conference on Computer Vision and Pattern Recognition}, pages 4724--4732, 2016.

\bibitem{weinreb2023keypoint}
Caleb Weinreb, Jonah Pearl, Sherry Lin, Mohammed Abdal~Monium Osman, Libby Zhang, Sidharth Annapragada, Eli Conlin, Red Hoffman, Sofia Makowska, Winthrop~F Gillis, et~al.
\newblock Keypoint-moseq: parsing behavior by linking point tracking to pose dynamics.
\newblock {\em BioRxiv}, 2023.

\bibitem{wiltschko2015mapping}
Alexander~B Wiltschko, Matthew~J Johnson, Giuliano Iurilli, Ralph~E Peterson, Jesse~M Katon, Stan~L Pashkovski, Victoria~E Abraira, Ryan~P Adams, and Sandeep~Robert Datta.
\newblock Mapping sub-second structure in mouse behavior.
\newblock {\em Neuron}, 88(6):1121--1135, 2015.

\bibitem{wiltschko2020revealing}
Alexander~B Wiltschko, Tatsuya Tsukahara, Ayman Zeine, Rockwell Anyoha, Winthrop~F Gillis, Jeffrey~E Markowitz, Ralph~E Peterson, Jesse Katon, Matthew~J Johnson, and Sandeep~Robert Datta.
\newblock Revealing the structure of pharmacobehavioral space through motion sequencing.
\newblock {\em Nature Neuroscience}, 23(11):1433--1443, 2020.

\bibitem{wiltshire2023deepwild}
Charlotte Wiltshire, James Lewis-Cheetham, Viola Komedov{\'a}, Tetsuro Matsuzawa, Kirsty~E Graham, and Catherine Hobaiter.
\newblock Deepwild: Application of the pose estimation tool deeplabcut for behaviour tracking in wild chimpanzees and bonobos.
\newblock {\em Journal of Animal Ecology}, 92(8):1560--1574, 2023.

\bibitem{yang2023unipose}
Jie Yang, Ailing Zeng, Ruimao Zhang, and Lei Zhang.
\newblock Unipose: Detecting any keypoints.
\newblock {\em arXiv preprint arXiv:2310.08530}, 2023.

\bibitem{yue2014learning}
Yisong Yue, Patrick Lucey, Peter Carr, Alina Bialkowski, and Iain Matthews.
\newblock Learning fine-grained spatial models for dynamic sports play prediction.
\newblock In {\em 2014 IEEE international conference on data mining}, pages 670--679. IEEE, 2014.

\bibitem{zhang2018unsupervised}
Yuting Zhang, Yijie Guo, Yixin Jin, Yijun Luo, Zhiyuan He, and Honglak Lee.
\newblock Unsupervised discovery of object landmarks as structural representations.
\newblock In {\em Proceedings of the IEEE Conference on Computer Vision and Pattern Recognition}, pages 2694--2703, 2018.

\bibitem{ZhangKptDisc18}
Yuting Zhang, Yijie Guo, Yixin Jin, Yijun Luo, Zhiyuan He, and Honglak Lee.
\newblock Unsupervised discovery of object landmarks as structural representations.
\newblock In {\em {IEEE} Conference on Computer Vision and Pattern Recognition, {CVPR}}, 2018.

\end{thebibliography}
}

\end{document}


\title{Supplement}

\maketitle

\section{Additional Experimental Results}~\label{sec:addtional_results}

\section{Additional Implementation Details}~\label{sec:additional_implementation}
\textbf{Architecture Details.} Our method uses ResNet-50~\cite{He2016DeepRL} as an encoder $\Phi$, GlobalNet~\cite{CPN17} as a pose decoder $\Psi$, and a series of convolution blocks as a reconstruction decoder $\psi$, following the unsupervised keypoint discovery model from~\cite{ryou2021weakly}. Architecture details about reconstruction decoder is shown in Table~\ref{tab:decoder_details}. 

\begin{table*}
    \caption{\textbf{Architecture details of the reconstruction decoder.} ``Conv\_block" refers to a basic convolution block which consists of 3$\times$3 convolution, batch normalization, and ReLU activation. }
    \label{tab:decoder_details}
    \begin{center}
        \begin{tabular}{lccc}
        \toprule
            Type & Input dimension & Output dimension & Output size  \\
            \midrule
            Upsampling & - & - &  16x16 \\
            Conv\_block & 2048 + \# keypoints $\times$ 2 $\times$ agents & 1024 & 16x16 \\
            Upsampling & - & - & 32x32 \\
            Conv\_block & 1024 + \# keypoints $\times$ 2 $\times$ agents & 512 & 32x32  \\
            Upsampling & - & - & 64x64 \\
            Conv\_block & 512 + \# keypoints $\times$ 2 $\times$ agents & 256 & 64x64  \\
            Upsampling & - & - & 128x128 \\
            Conv\_block & 256 + \# keypoints $\times$ 2 $\times$ agents & 128 & 128x128  \\
            Upsampling & - & - & 256x256 \\
            Conv\_block & 128 + \# keypoints $\times$ 2 $\times$ agents & 64 & 256x256 \\
            Convolution & 64 & 3 & 256x256 \\
            \bottomrule\\[-1em]
        \end{tabular}
    \end{center}
\end{table*}

\textbf{Hyperparameters.} The hyperparameters for the keypoint discovery model are listed in Table~\ref{tab:hyperparameter_1}. Unless otherwise specified, all models use the SSIM image as the reconstruction target. Each keypoint discovery model is trained on an NVIDIA A100 Tensor Core GPU until the training loss converges. 

\begin{table*}[!t]
  \centering
  \small
  \scalebox{1.0}{
   \begin{tabular}{c | c | c| c | c| c | c } 
   \hline
   Dataset & \# Keypoints & Batch size & Resolution & Frame Gap & Learning Rate & Agents \\
   \hline
    CalMS21 & 10 &  5 & 256 & 6 & 0.001 & 2 \\
    \hline
    Fly vs. Fly & 10 & 5 & 256 & 3 & 0.001 & 2 \\
   \hline
    PAIR-R24M & 12 & 5 & 256 & 6 & 0.001 & 2 \\
   \hline  
   \label{tab:keypoint_details}
\end{tabular}
}
\caption{{\bf Hyperparameters for Keypoint Discovery.}} \label{tab:hyperparameter_1}
\end{table*}

\textbf{Segmentation Parameters.} Our method utilizes the DEVA framework \cite{DEVA} for video object segmentation, guided by text prompts. The parameters for DEVA segmentation are detailed in Table~\ref{tab:segmentation_details}. In text-prompted mode, we use the following key parameters:

\begin{itemize}
\item \textbf{DINO\_THRESHOLD}: This is the threshold for DINO to consider a detection as valid. A higher threshold would result in fewer but more confident detections, while a lower threshold might include more detections but with a risk of false positives.
  
\item \textbf{prompt}: This is the text prompt used to guide the segmentation. The prompts are separated by a full stop; for example, "rat.mouse". The wording of the prompt and minor details like pluralization might affect the results. For instance, "cats" might yield different results from "cat".

\item \textbf{size}: This is the internal processing resolution for the propagation module, which defaults to 480. A higher resolution might lead to more detailed segmentations but would require more computational resources.
\end{itemize}

All other parameters were kept at their default values.

\begin{table*}[!t]
  \centering
  \small
  \scalebox{1.0}{
   \begin{tabular}{c | c | c | c } 
   \hline
   Dataset & Grounded Segmentation Prompt & DINO Threshold & Size  \\
   \hline
    CalMS21 & rat.mouse & 0.45 & 480 \\
    \hline
    Fly vs. Fly & fly & 0.5 & 480 \\
   \hline
    PAIR-R24M & rat.mouse & 0.45 & 480 \\
   \hline   
   \label{tab:segmentation_details}
\end{tabular}
}
\caption{{\bf Parameters for Segmentation.}} \label{tab:parameter_1}
\end{table*}

\section{Added computational cost}

\begin{figure}[htbp]
    \centering
    \includegraphics[clip, width=0.25\textwidth]{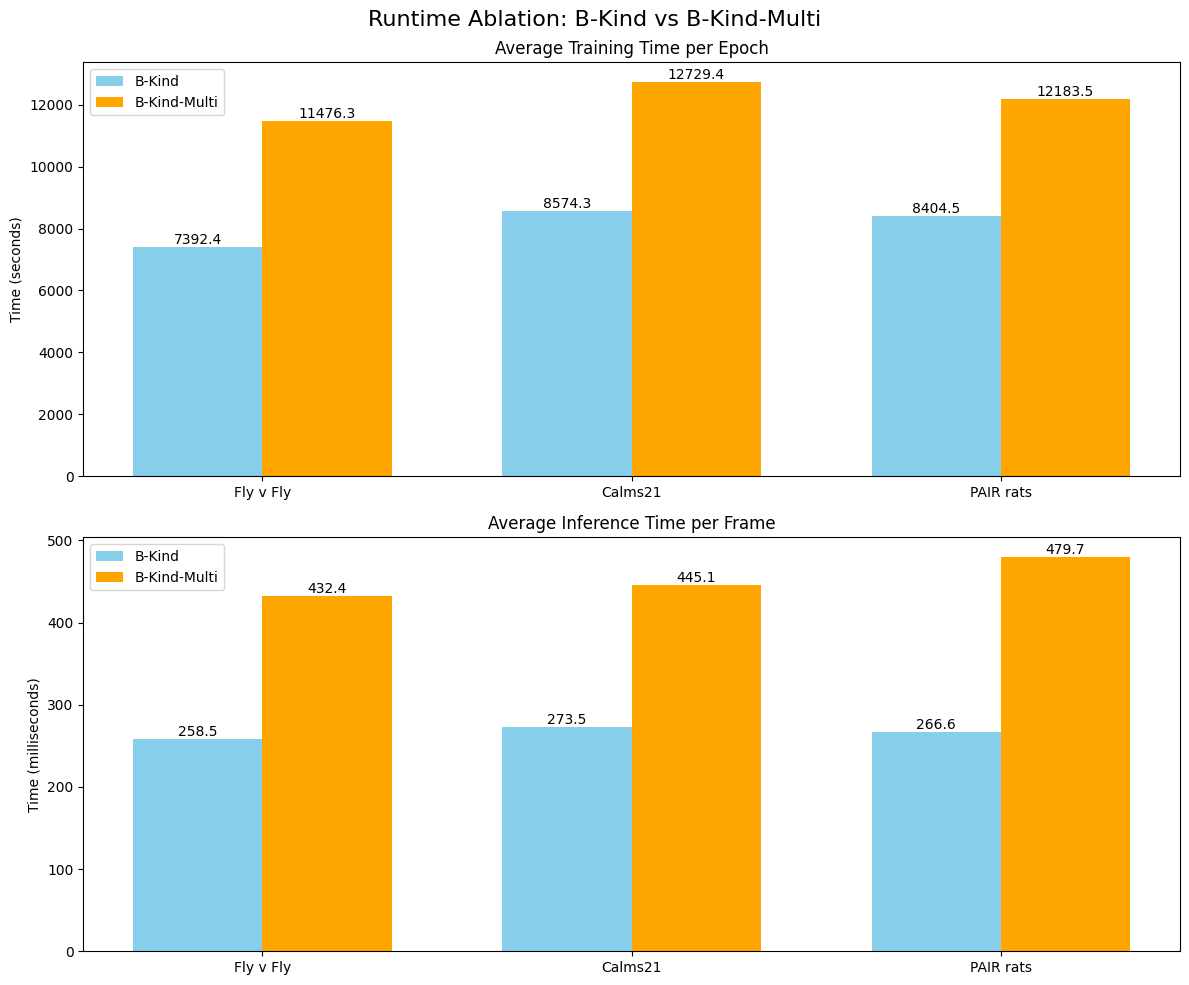}
    \caption{Computational Overhead comparison of B-Kind and B-Kind Multi on different datasets. The training time for B-Kind-Multi is only about 40\% longer in both training and inference times.}
    \label{fig:placeholder}
\end{figure}

\section{Additional Ablations}

\begin{table}
  \begin{center}
  \scalebox{0.7}{
    \begin{tabular}{lc}
        \toprule[0.2em]
        Video Segmentation Type &  \%-MSE \\
        \toprule[0.2em]
        \multicolumn{2}{c}{\textbf{\textit{CalMS21}}} \\
        Frame by Frame & \textbf{3.454 $\pm$ .167} \\
        DEVA & \textbf{.693 $\pm$ .061} \\
        \bottomrule[0.1em]
        \multicolumn{2}{c}{\textbf{\textit{Fly v. Fly}}} \\
        Frame by Frame & \textbf{2.531 $\pm$ .097} \\
        DEVA & \textbf{.115 $\pm$ .032} \\
        \bottomrule[0.1em]
        \multicolumn{2}{c}{\textbf{\textit{PAIR-R24M}}} \\
        Frame by Frame & \textbf{2.874 $\pm$ .121} \\
        DEVA & \textbf{.168 $\pm$ .027} \\
        \bottomrule[0.1em]
    \end{tabular}}
  \caption{\textbf{Video segmentation ablation on CalMS21, Fly, and PAIRS}. \%-MSE error is reported by changing the segmentation type. Extracted features correspond to keypoint locations, confidence, and covariance. Results are from 5 B-KinD runs.}
  \label{tab:ablation_segmentation}
  \end{center}\vspace{-0.6cm}
\end{table}

\begin{table}
  \begin{center}
  \scalebox{0.7}{
    \begin{tabular}{cccc}
        \toprule[0.2em]
        Recon. & Rot. Eq. & Sep. & \%-MSE \\
        \toprule[0.2em]
        \multicolumn{4}{c}{\textbf{\textit{Fly v. Fly}}} \\
        \checkmark & & & $.298 \pm .067$ \\
        \checkmark & \checkmark & & $.173 \pm .012$ \\
        \checkmark & & \checkmark & $.221 \pm .059$ \\
        \checkmark & \checkmark & \checkmark & $.115 \pm .032$ \\
        \bottomrule[0.1em]
        \multicolumn{4}{c}{\textbf{\textit{CalMS21}}} \\
        \checkmark & & & $.896 \pm .021$ \\
        \checkmark & \checkmark & & $.722 \pm .031$ \\
        \checkmark & & \checkmark & $.764 \pm .055$ \\
        \checkmark & \checkmark & \checkmark & $.693 \pm .061$ \\
        \bottomrule[0.1em]
        \multicolumn{4}{c}{\textbf{\textit{PAIR-R24M}}} \\
        \checkmark & & & $.248 \pm .015$ \\
        \checkmark & \checkmark & & $.192 \pm .037$ \\
        \checkmark & & \checkmark & $.247 \pm .065$ \\
        \checkmark & \checkmark & \checkmark & $.145 \pm .056$ \\
        \bottomrule[0.1em]
    \end{tabular}}
  \caption{\textbf{Loss Ablation Results on Fly v. Fly, CalMS21, and PAIR-R24M}. ``Recon." represents reconstruction loss, ``Rot. Eq." represents rotation equivariance loss, and ``Sep." represents separation loss. \%-MSE error is reported for different combinations of loss functions. Results are from 5 runs.}
  \label{tab:ablation_loss}
  \end{center}\vspace{-0.4cm}
\end{table}

\begin{table}
  \begin{center}
  \scalebox{0.7}{
    \begin{tabular}{lc}
        \toprule[0.2em]
        Image Resolution &  \%-MSE \\
        \toprule[0.2em]
        \multicolumn{2}{c}{\textbf{\textit{Fly v. Fly}}} \\
        128x128 & \textbf{.129 $\pm$ .023} \\
        256x256 & \textbf{.115 $\pm$ .032} \\
        512x512 & \textbf{.110 $\pm$ .038} \\
        \bottomrule[0.1em]
        \multicolumn{2}{c}{\textbf{\textit{CalMS21}}} \\
        128x128 & \textbf{.747 $\pm$ .059} \\
        256x256 & \textbf{.693 $\pm$ .061} \\
        512x512 & \textbf{.603 $\pm$ .065} \\
        \bottomrule[0.1em]
        \multicolumn{2}{c}{\textbf{\textit{PAIR-R24M}}} \\
        128x128 & \textbf{.180 $\pm$ .031} \\
        256x256 & \textbf{.168 $\pm$ .027} \\
        512x512 & \textbf{.145 $\pm$ .056} \\
        \bottomrule[0.1em]
    \end{tabular}}
  \caption{\textbf{Image resolution ablation on Fly v. Fly, CalMS21, and PAIR-R24M}. \%-MSE error is reported for different image resolutions. Results are from 5 B-KinD runs.}
  \label{tab:ablation_resolution}
  \end{center}\vspace{-0.6cm}
\end{table}

{\small
\bibliographystyle{ieee_fullname}
\bibliography{egbib}
}